\documentclass[conference]{IEEEtran}
\IEEEoverridecommandlockouts
\usepackage{cite}
\usepackage{amsmath,amssymb,amsfonts}

\usepackage{graphicx}
\usepackage{textcomp}
\usepackage{xcolor}
\def\BibTeX{{\rm B\kern-.05em{\sc i\kern-.025em b}\kern-.08em
    T\kern-.1667em\lower.7ex\hbox{E}\kern-.125emX}}
    
\usepackage{hyperref}
\usepackage{amsmath}
\usepackage{wrapfig,lipsum,booktabs}
\usepackage{floatrow}
\usepackage{subcaption}
\DeclareMathOperator*{\argmax}{arg\,max}
\usepackage[
separate-uncertainty = true,
multi-part-units = repeat
]{siunitx}
\usepackage{multirow}
\usepackage{bbding}
\usepackage{algorithm}
\usepackage{algorithmic}

\begin{document}

\title{Node Embedding using Mutual Information and Self-Supervision based Bi-level Aggregation\\
}

\author{\IEEEauthorblockN{Kashob Kumar Roy$^{\dagger}$, Amit Roy$^{\dagger}$\thanks{$^\dagger$ Equal Contribution}, A K M Mahbubur Rahman, M Ashraful Amin and Amin Ahsan Ali}
\IEEEauthorblockA{\textit{Artificial Intelligence and Cybernetics Lab},
\textit{Independent University Bangladesh}\\
\{kashobroy, amitroy7781\}@gmail.com, and \{akmmrahman, aminmdashraful, aminali\}@iub.edu.bd}}


\maketitle

\begin{abstract}
Graph Neural Networks (GNNs) learn low dimensional representations of nodes by aggregating information from their neighborhood in graphs. However, traditional GNNs suffer from two fundamental shortcomings due to their local ($l$-hop neighborhood)  aggregation scheme. First, not all nodes in the neighborhood carry relevant information for the target node. Since GNNs do not exclude noisy nodes in their neighborhood, irrelevant information gets aggregated, which reduces the quality of the representation. Second, traditional GNNs also fail to capture long-range non-local dependencies between nodes. To address these limitations, we exploit mutual information (MI) to define two types of neighborhood, 1) \textit{Local Neighborhood} where nodes are densely connected within a community and each node would share higher MI with its neighbors,  and 2) \textit{Non-Local Neighborhood} where  MI-based node clustering is introduced to assemble informative but graphically distant nodes in the same cluster. To generate node presentations, we combine the embeddings generated by bi-level aggregation - local aggregation to aggregate features from local neighborhoods to avoid noisy information and non-local aggregation to aggregate features from non-local neighborhoods. Furthermore, we leverage self-supervision learning to estimate MI with few labeled data. Finally, we show that our model significantly outperforms the state-of-the-art methods in a wide range of assortative and disassortative graphs\footnote{Source Code at: \href{https://github.com/forkkr/LnL-GNN}{\color{blue}{https://github.com/forkkr/LnL-GNN}}}.
\end{abstract}


\section{Introduction}
In recent years, Graph Neural Networks (GNNs) have seen a tremendous amount of success in formulating a variety of real-world applications and providing low dimensional high-level representations or embeddings of nodes and graphs. These latent representations are useful and effective in several node-based downstream machine learning tasks such as node classification, link prediction, traffic forecasting, etc., and graph-based tasks like graph classification, graph reconstruction, human activity recognition, etc. The most popular message-passing GNNs include~\cite{hamilton2017inductive,kipf2016semi,velivckovic2017graph,velickovic2019deep,xu2018representation} that tend to find node embeddings by aggregating information from the local neighborhood in different approaches.

\begin{figure}[tb]
{\caption{Two weaknesses of classic GNNs: \textit{i)} incompetent to distinguish between relevant (enclosed by green boundary) and irrelevant nodes in $l$-hop local-neighborhood; \textit{ii)} less effective in capturing feature information from distant but similar nodes (two green center points).}
\label{fig:graph_Motivation}}
{\includegraphics[width=\columnwidth]{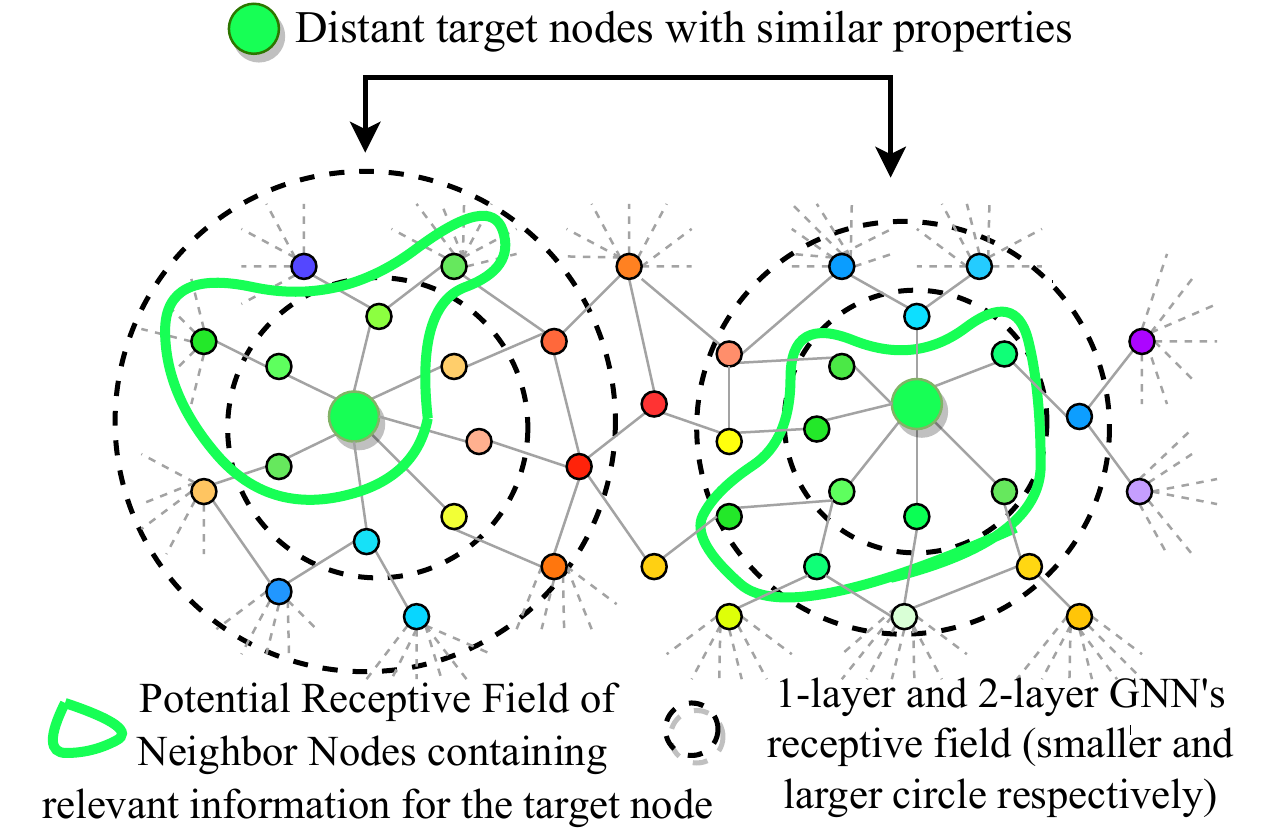}}
\end{figure}
However, the main objective of GNNs in node classification is that the learned embeddings of nodes with the same class labels should be close to each other in the latent space, and there exists a large separation among the embeddings of different classes. GNNs with their multi-layer local aggregation framework, tend to aggregate features from the proximal nodes with the assumption that the nodes in proximity are of similar properties. As shown in Fig.~\ref{fig:graph_Motivation}, GNN with $l$-layers can capture information from $l$-hop neighborhood~\cite{kipf2016semi}, where the increasing depth of GNNs expands the receptive field of feature information. The local aggregation strategy is more suited for node classification in assortative graphs where the homophily assumption holds. That is when nodes in proximity share the same class label. However, existing GNNs fail to distinguish between relevant and noisy nodes in their local-neighborhood. Moreover, in disassortative graphs, many nodes of the local neighborhood do not have the same class label and thus carry irrelevant information for this node. This leads to the incorporation of noise and irrelevant information during local aggregation and reduces the quality of the representations produced. 


Another major drawback of capturing information only from the local neighborhood is that GNNs fail to incorporate information from distant but informative nodes in the graph. Distant nodes with the same class label that show high structural similarity, e.g., similar degree distribution in their neighborhood or similar role in the network, etc., may carry relevant information for each other~\cite{pei2020geom}. 
A few recent works~\cite{xu2018representation,li2019deepgcns,rong2019dropedge} try to increase the receptive field of GNNs by increasing the number of layers. However, GNNs with multi-layered architecture suffer from over-smoothing problem~\cite{chen2020measuring,li2018deeper}, where relevant information gets mixed up with noise. Recently, GEOM-GCN\cite{pei2020geom} attempts to capture long-range dependencies by employing a bi-level aggregation that accumulates information from both graphical neighborhood and latent space neighborhood, which is defined by the euclidean distance of pre-trained embeddings. NL-GNN\cite{liu2020non} leverages attention-guided sorting with a single calibration vector to define non-local neighborhood. Still,
the non-local neighborhood defined by euclidean distance or single calibration vector can not explicitly capture the long-range non-linear dependencies between nodes.

Unlike existing models, in this work, we consider both graph topology and feature-based correlations between nodes to define the local neighborhood or community of a node. The local neighborhood of a target node should consist of nodes that are densely connected to the target node and correlated in their feature space. It promotes our model to avoid noisy nodes and draw the effective data-specific receptive field of information for nodes. We integrate graphical connectivity and feature correlations into the existing community detection method~\cite{jin2021survey} to identify the intrinsic communities as local neighborhoods. Here, Mutual Information (MI) is used to encode the non-linear correlations between two nodes' features as edge weights because of its superior ability to capture non-linear relationships between nodes compared to euclidean distance, or cosine similarity, or learnable attention measures.
Consequently, the defined local neighborhood enables our model to overcome the over-smoothing problem to some extent through aggregating features from the relevant nodes. The proposed method does not need to set a fixed value of $l$ to define the receptive field. However, the community detection algorithm can not bring distant informative nodes to the local neighborhood. Thus, we utilize MI to determine non-local neighborhood as a group of nodes that are highly correlated in their latent space but not local according to graph topology. The proposed method performs the bi-level aggregation - local aggregation among local neighbor nodes and non-local aggregation among non-local neighbors. It thus can generate distinguishable and better informative embeddings for nodes.  

However, MI estimation is quite challenging if there are only a few labeled data points. Therefore, we propose a novel training approach to estimate MI by leveraging a self-supervised approach\cite{you2020does} to utilize unlabeled data.

In summary, our key contributions are as following-
\begin{enumerate}
    \item We propose an efficient MI estimator that leverages self-supervised learning approach to improve the training even when only a few nodes are given labels.
    \item We exploit both MI and community detection together to define the hyper-parameter free data-specific local neighborhood. It allows our model to capture community structural information and leverage important features from multi-hop neighboring nodes without being susceptible to noise.
    \item We propose MI guided non-local neighborhood that enables our model to aggregate features from distant correlated nodes to learn informative representations.
    \item We introduce a bi-level aggregation scheme, which is superior in both assortative and disassortative graphs.
    \item We have conducted extensive experiments on widely used ten graph datasets to verify the efficacy of our model and observed that it outperforms the state-of-the-art approaches by a good margin. 
\end{enumerate}
\section{Background and Related Works}
\label{sec:background_and_related_works}
We consider a graph $\mathcal{G} = (V, E)$ where $V$ and $E$ are the set of nodes and edges respectively. Each edge $e \in E$ connects a node pair, $ (u, v) \in (V \times V)$. The $l$-hop neighborhood of a node $u$ refers to a set of nodes that are reachable from $u$ within $l$ edges. Each node $u \in V$ has a $d$-dimensional feature vector $X_u \in R^{d}$. Besides, some nodes have class labels from a defined set of labels, $L = \{1,..,c\}$. The goal of node classification task is to learn a mapping, $ \mathcal{M}: V \longrightarrow L$ and predict the labels of unlabeled nodes.

\subsection{Homophily Ratio, HR} Node homophily of a graph $\mathcal{G}$  refers to how likely nodes with the same class label are near each other in the graph. The node homophily is defined in~\cite{pei2020geom} as 
$HR(\mathcal{G}) = \frac{1}{|V|}\sum_{u \in V}\frac{Ns_{u}}{Ng_{u}}$
where, $Ns_{u}$ and $Ng_{u}$ are the number of direct neighbors that have the same class label as $u$ and the total numbers of direct neighbors of $u$ respectively. $|V|$ is the total number of nodes.
\subsection{Assortative and Disassortative Graphs}
Assortative graphs have high node homophily, while low homophily is observed in disassortative graphs. Low node homophily means nodes with the same class labels are more likely to be distant from each other. Graph datasets for the node classification tasks are categorized based on their homophily ratio that has been shown in Table~\ref{tab:dataset_description}.
\subsection{Graph Neural Networks (GNNs)} 
Among recent popular GNNs, GCN~\cite{kipf2016semi} proposes spectral graph convolution. GraphSAGE~\cite{hamilton2017inductive} introduces neighborhood sampling and aggregation based inductive framework for large graphs. GAT~\cite{velivckovic2017graph} first adapts attention mechanism into feature aggregation. But all of them suffer from  over-smoothing problem while performing feature aggregation in more layers. JK Network~\cite{xu2018representation}, DNA~\cite{fey2019just} enables deeper GNNs by introducing jump connection and task-specific receptive fields in computing embeddings, respectively. 
Although these methods perform well for assortative graphs, feature aggregation only from the local neighborhood is not enough to take care of non-local dependencies among nodes, especially in disassortative graphs. GEOM-GCN~\cite{pei2020geom} proposes a virtual non-local neighborhood-based aggregation scheme based on the euclidean distance among nodes in embedding space. 
But the use of non-task-specific pre-trained node embeddings and the lack of ability to capture non-linear relationships limit the effectiveness of~\cite{pei2020geom}. 
NL-GNN~\cite{liu2020non} defines non-local neighborhood based on attention-guided sorting with a single calibration vector. 
\subsection{Community Detection and MI:} Traditionally, community detection algorithms find communities by maximizing some metric such as modularity~\cite{newman2006modularity} that represents the quality of a community. M-NMF~\cite{wang2017community} attempts to preserve community structures in hidden representations. Jin et al.~\cite{jin2021survey} demonstrate several advancements of integrating topology and node feature into community detection methods. Hence, MI can be exploited to find communities as it is a measure of the mutual dependence between two random variables. The estimation of MI using neural networks has gained much attention in recent years. MINE~\cite{belghazi2018mutual} proposes a neural MI estimator by exploiting gradient descent over neural networks. DIM~\cite{hjelm2018learning} introduces methods for estimating and maximizing MI between local and global representations for image data, whereas DGI~\cite{velickovic2019deep}, MI-StrutRL~\cite{ijcai2020-472} are proposed to optimize MI between node-level representation and graph-level summary representation.

\begin{figure*}[thb]
    \centering
    \includegraphics[width=0.8\textwidth]{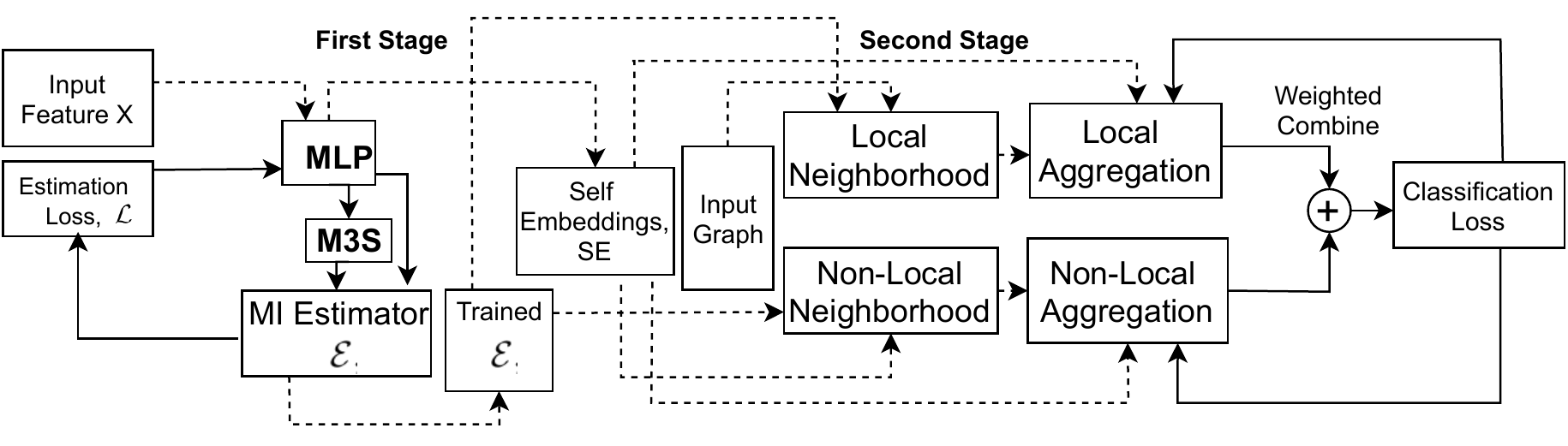}
    \caption{Flow chart of LnL-GNN. At the first stage, MLP, to compute SEs, and MI Estimator, $\mathcal{E}$, are trained using the estimation loss $\mathcal{L}$. At next stage, obtained SEs and trained $\mathcal{E}$ are used to define local and non-local neighborhood followed by performing the local and non-local aggregation to compute the final embeddings of nodes. And, it updates the parameters of aggregation modules by optimizing the classification cross-entropy loss.  Here, M3S denotes Multi-stage Self-Supervised Sampling. Dashed Line(- - -): Pre-processed Input Propagation, Solid Line (---) : Weight Update.}
    \label{fig:model_diagram}
\end{figure*}

\section{Proposed Model}
In this section, we will describe our proposed model (LnL-GNN) as shown in Fig.~\ref{fig:model_diagram}. At first stage, self embeddings (SE) are generated for each node from the features of the nodes. Also, an MI estimator is trained, which computes the MI between two nodes using the SEs of the nodes. These are used in the second state, where local and non-local neighborhoods are computed, and bi-level aggregation is performed. These steps are described in detail below.  

\subsection{Self-Embedding, SE:} 
We use a simple multilayer perceptron, $\mathcal{Z}_u=MLP(X_u)$ with SELU activation function to extract low-dimensional representations named as self-embeddings (SEs) from only ego (target) node features without aggregating features from the neighborhood. Except for the graphs with a high homophily ratio, we can utilize the mean-aggregated features of 1-hop neighbors that we will discuss later in the experimental section.

\subsection{MI Estimator, MIE:} MI has the capability of capturing non-linear dependencies between node representations to distinguish nodes of different classes. Intuitively, MI between the feature vectors of node pairs having the same class label should be higher than that for node pairs with different class labels. As DGI~\cite{velickovic2019deep} and MI-StrutRL~\cite{ijcai2020-472} compute MI between node representation and graph-level summary representation, we can not use them directly here. Therefore, we define a novel MI estimator MIE to compute MI between two node representations, as follows:
\begin{equation}
    \mathcal{E}(\mathcal{Z}_{u},\mathcal{Z}_{v}) \to \alpha(\mathcal{Z}_{u}^{T}W_{e}\mathcal{Z}_{v} + b)
\end{equation}
where $\mathcal{Z}_{u}$ and $\mathcal{Z}_{v}$ are self-embeddings of node $u$ and $v$ respectively, and $W_{e}$ and $b$ are learnable weight and bias parameters. $\alpha$ is a non-linear activation function. 

Recent research~\cite{hjelm2018learning,velickovic2019deep} showed that a noise-contrastive objective function with a standard binary-cross entropy loss between the positive and negative samples could effectively maximize and estimate MI between two node representations. Following their works, we define our estimation loss function as,
\begin{equation}
\mathcal{L}=-\sum_{u \in V} \sum_{v, \tilde{v}} E_{v}[\log \mathcal{E}(\mathcal{Z}_{u},\mathcal{Z}_{v})]-E_{\tilde{v}}[\log(1-\mathcal{E}(\mathcal{Z}_{u},\mathcal{Z}_{\tilde{v}}))]
\label{eq:mi_loss}
\end{equation}
where, $v$ and $\tilde{v}$ denote positive and negative samples respectively. For a node $u$, nodes in the graph with the same class label as $u$ are considered as positive samples and nodes with different class labels as negative samples.

\begin{figure}[thb]

\includegraphics[width=\columnwidth]{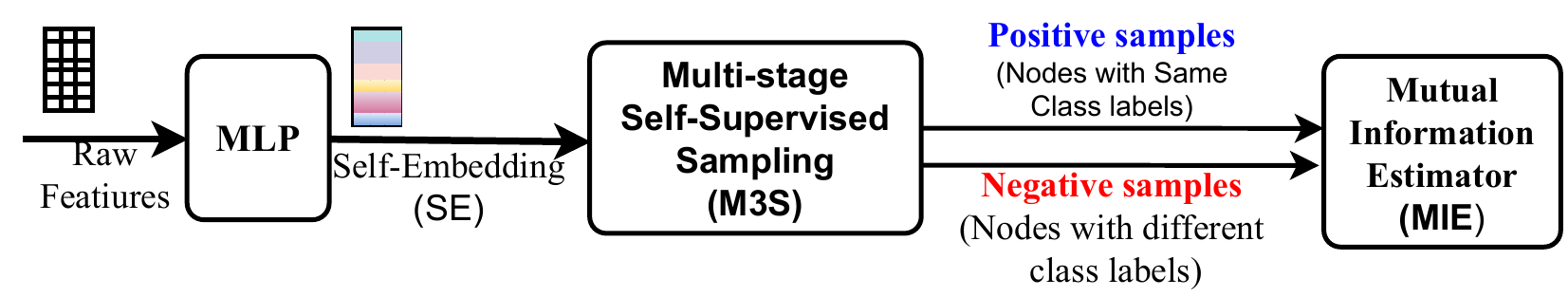}
\caption{MLP embeds node features into SEs.  M3S assign pseudo-labels to unlabeled nodes and include them into the set of labeled nodes. Next it draws positive and negative samples. MIE attempts to estimate MI between these samples. Finally, the parameters will be learned through optimizing the loss function in Eq.~\ref{eq:mi_loss}.}
\label{fig:mi_se_fig}
\end{figure}

\subsection{Multi-stage Self-Supervised Sampling, M3S:} With few labeled data, it is very challenging to optimize the MIE well.
We introduce a novel multi-stage self-supervised sampling (M3S) to utilize the unlabeled data in training the MIE as shown in Algortihm~\ref{algo:m3s}.

\begin{algorithm}[H]
  \caption{Multi-stage Self-Supervised Sampling, M3S}
  \label{algo:m3s}
  \begin{flushleft}
    \textbf{INPUT:} Input feature $X$ \\
    \textbf{OUTPUT:} Estimator $\mathcal{E}$, Self-Embedding $\mathcal{Z}$
    \end{flushleft}
  \begin{algorithmic}[1]
    \STATE \textbf{function} M3S
    \STATE $\mathcal{E}, \mathcal{Z} \gets \text{train MIE for a fixed number of epochs}$
    \FOR{each stage}
    \STATE Compute centroids of each class
    \FOR{each unlabeled node, u}
    \STATE Compute MI between u and each centroid
    \STATE Assign pseudo-class label to u
    \ENDFOR
    \STATE Sort all pseudo-labeled nodes based on MI
    \STATE Select top $t$ nodes with higher MI
    \STATE Add t nodes to labeled data
    \STATE \small $\mathcal{E}, \mathcal{Z} \gets \text{train MIE on updated labeled data}$
    \ENDFOR
    \STATE \textbf{return} $\mathcal{E}, \mathcal{Z}$
    \STATE \textbf{end function}
  \end{algorithmic}
\end{algorithm}

Initially, we train MIE using the positive and negative samples drawn from labeled nodes for a fixed number of epochs to obtain meaningful SEs (see in Line 2s). Afterward, we compute the centroids of each class of labeled nodes and compute MI between class centroids and self-embeddings of unlabeled nodes. Next, we assign the class label with maximum MI to each unlabeled node as their pseudo-label. We sort the unlabeled nodes in descending order based on the computed MI values. Then, we select top $t$ unlabeled nodes,  add them into the labeled data for the next epoch of training and remove them from unlabeled data. This whole M3S process will be repeated several times until MIE to be well optimized. Fig.~\ref{fig:mi_se_fig} shows the overview of the whole process to learn SEs along with MIE by optimizing the estimation loss, $\mathcal{L}$ in Eq.~\ref{eq:mi_loss}.


Following the formulation of~\cite{ijcai2020-472,hjelm2018learning}, we define MI between two nodes $u$ and $v$ approximately as:
\begin{equation}
MI(u,v) = E_{v}[-sp(-\mathcal{E}(\mathcal{Z}_{u},\mathcal{Z}_{v}))] + E_{\widetilde{v}}[sp(\mathcal{E}(\mathcal{Z}_{u},\mathcal{Z}_{v}))]
\label{eqn:mi}
\end{equation}
where $sp$ is Softplus function, $sp(c) = \log(1+exp(c))$. It should be noted that Equation~\ref{eqn:mi} actually defines a new class of information measures. Nevertheless, the expressive power of neural network ensures that the esitmator $\mathcal{E}$ can approximate the mutual information with an arbitrary accuracy.


\begin{figure*}[t]
\centering
\includegraphics[width=\textwidth]{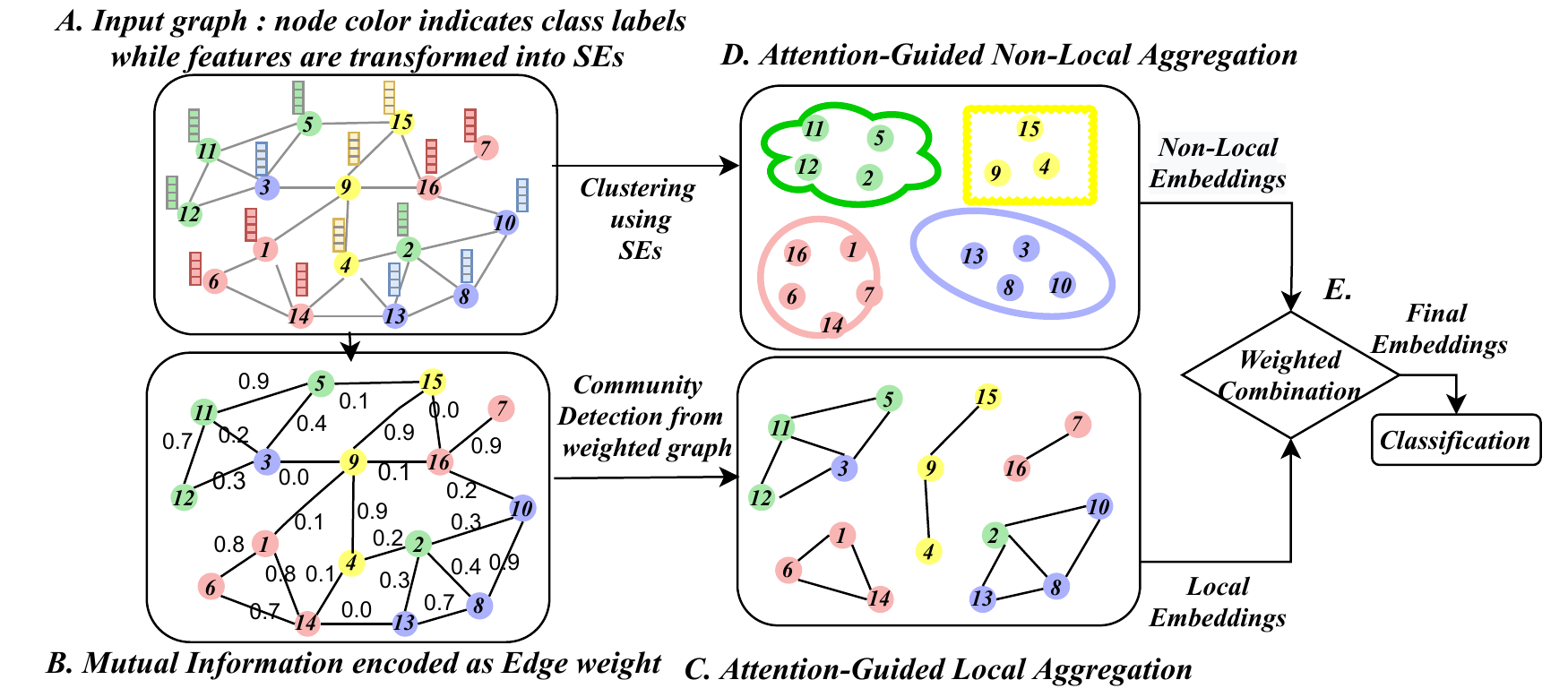}
\caption{\textbf{\textit{A.}}  The features of all nodes are transformed into SEs using an MLP depicted in Fig.~\ref{fig:mi_se_fig}. \textbf{\textit{B-C.}} Local Embedding: MI between the SEs of adjacent nodes is encoded as edge weight in the graph. Then, the weighted graph is processed to identify local neighborhood, and then local aggregation is performed to compute node embeddings. \textbf{\textit{D.}} Non-Local Embedding: MI based clustering is applied on SEs of nodes to find non-local neighborhood, followed by the non-local aggregation to compute node embeddings. \textbf{\textit{E.}} Finally, local-embeddings and non-local embeddings are combined and transformed to the final embeddings that are used for node classification, and parameters are learned by optimizing classification loss.}
\label{fig:le_nle_fig}
\end{figure*}

\subsection{Bi-level Neighborhood Scheme}
To overcome the drawbacks of traditional GNNs, as mentioned earlier, we introduce local neighborhood to aggregate features from local relevant nodes while ignoring the noisy nodes and non-local neighborhood to aggregate information from distant important nodes. We determine the local neighborhood by integrating community detection from the graph and MI (node features-based correlation). MI between two adjacent node features is encoded as edge weight of the graph, and then community detection is performed on this weighted graph. The resulting communities define a local neighborhood for the nodes (see in Fig.~\ref{fig:le_nle_fig}(B-C)). Intuitively, aggregating features from nodes within the community rather than from all $l$-hop neighbors is better as nodes within the community are densely connected to each other than to the rest of the graph, and MI helps to eliminate possibly noisy nodes (nodes with different class labels) from the local neighborhood. We also perform clustering on node features while ignoring the graph topology. This puts nodes, which are topologically non-local but has similar features, in the same cluster. Thus aggregating both from the local neighborhood (graph community) and non-local neighborhood (nodes in the same cluster) can produce better embeddings for node classification.
\subsubsection{Local Neighborhood through Community Detection}
Steps of delineating local neighborhood in the graph are as follows:
\begin{itemize}
\item Edge Weight Assignment: We assign weights on edges as, $\forall e_{uv} \in E: W_{uv} = MI(u,v)$. But we do not alter any edges in the adjacency matrix and keep original structural information unchanged. We use this transformed weighted graph to find communities.
\item Weighted Modularity: Modularity is defined as the fraction of edges that fall within communities minus the expected value of the same quantity if edges are assigned at random, conditional on the given community
memberships and connectivities of nodes. For a particular partitioning of graph into communities, the modularity for a weighted graph can be formulated~\cite{newman2004analysis} as,
\begin{equation}
    \mathcal{Q}(P_{i}) = \frac{1}{2|W|} \sum_{u, v} [W_{uv}-\frac{S(u)S(v)}{2|W|}]\delta_{uv}
\end{equation}
where, $S(u)$ is the sum of weights of edges associated with node $u$, $\delta_{uv}$ is 1 when both nodes $u$ and $v$ are in the same community under partitioning $P_{i}$ otherwise 0, and $|W|$ is the sum of weights of all edges. The value of $\mathcal{Q}$ could be positive or negative where the larger positive value of $\mathcal{Q}$ indicates better partitioning of the graph into communities.
\item Community Detection:
We use Louvain algorithm~\cite{blondel2008fast} which looks for a set of communities in the graph with maximum modularity in bottom-up fashion starting each node as a single community and merge different communities such that weighted modularity is maximized. Once obtained the optimal partitioning $P^{*}$, we define local neighborhood, $\mathcal{N_{L}}$ as following,
\begin{equation}
\begin{split}
    &P^{*} = \argmax_{P_{i} \in \mathcal{P}}\mathcal{Q}(P_{i})\\
    &\mathcal{N_{L}}(u) = \{\forall v \in V: \delta_{v} == \delta_{u}\}
\end{split}
\end{equation}
where the set of possible partitioning, $\mathcal{P}$$=\{P_i,..,P_{m}\}$, ${P}_{i}$$ =\{p_{1},..,p_{k}\}$, $p_{i}$ is a single community, and $\delta_{u}$ denotes community membership of $u$. 
\end{itemize}
After that, we use $\mathcal{N_{L}}$ to capture structural information and to aggregate features from local neighboring nodes.

\subsubsection{Non-Local Neighborhood}
We utilize MI to group nodes that are closely correlated in their self-embedding space together into the same cluster. Steps of MI based clustering are as follows -
\begin{itemize}
\item Initialization: It decomposes data-points randomly into a set of disjoint clusters where the number of clusters is defined before. Data points are the SEs of individual nodes.
\item Update assignment: It updates the assignment of a node, $u$ as following, where $\mathcal{C}$ is set of clusters,
\begin{equation}
    \kappa_{u} = \argmax_{c \in \mathcal{C}}\frac{1}{|c|}\sum_{v \in c} MI(\mathcal{Z}_{u},\mathcal{Z}_{v})
    \label{eq:mic}
\end{equation}
\item Termination: Similar to $K$-Means, it updates the assignments of all nodes iteratively until it reaches the maximum iteration limit or converges.
\end{itemize}

where $\mathcal{C}$ is set of clusters, $\mathcal{Z}_{u}$ and $\mathcal{Z}_{v}$ are self-embeddings of node $u$ and $v$.


\begin{table*}[t]
\begin{tabular}{c|ccccccc||ccc}
\hline
\multicolumn{8}{c||}{\textbf{Disassortative}} & \multicolumn{3}{c}{\textbf{Assortative}}                        \\ \hline
\textbf{Datasets -$>$} & \textit{Texas} & \textit{Cornell}  & \textit{Wisconsin} & \textit{Washington}  & \textit{Squirrel} & \textit{Actor} & \textit{Chameleon}  & \textit{Citeser} & \textit{Pubmed} & \textit{Cora}\\ \hline

\textbf{HR} & \textbf{0.06} & \textbf{0.11}& \textbf{0.16} &\textbf{0.16} & \textbf{0.22}& \textbf{0.24}& \textbf{0.25}  &\textbf{0.71}& \textbf{0.79} & \textbf{0.83}\\
\textbf{\# Nodes } & 183     & 183   & 251   &   230       & 5201     & 7600 & 2277     & 3327    & 19717  & 2708  \\ 
\textbf{\# Edges}  & 309 & 295        & 499    &  446       & 217073   & 33544 & 36101  & 4732    & 44338  & 5429 \\ 
\textbf{\# Avg. Deg.} & 3.05 & 3.03        & 3.59    &  3.18  & 76.27   & 7.02 & 27.55   & 2.74    & 4.50 & 3.90\\ 
\textbf{\# Features}   & 1703    & 1703  & 1703    &  1703   & 2089     & 931 & 2325   & 3703    & 500   & 1433\\ 
\textbf{\# Classes}       & 5         & 5        & 5     &     5    & 5     & 5      &   5  & 6       & 3 & 7    \\ \hline
\end{tabular}
\caption{Description of datasets used in our experiments}
\label{tab:dataset_description}
\end{table*}

Once the clusters of nodes are obtained, we define non-local neighborhood in Eq.~\ref{eq:nlong},
\begin{equation}
\mathcal{N_{\widetilde{L}}}(u) = \{\forall v \in V: \kappa_{v} == \kappa_{u}\}    
\label{eq:nlong}
\end{equation}
However, for large datasets, aggregating information from all nodes in non-local neighborhood would be computationally expensive with respect to time. To overcome this problem, we sample nodes based on their centrality measures against the whole graph. It helps to filter out structurally similar nodes to aggregate information.

As we mentioned above, the local neighborhood is usually different from the non-local neighborhood of a node. Our model gives more importance on local connectivity than feature-based similarity in local neighborhood, in contrast, non-local neighborhood depends solely on feature-based correlation. For example, in Fig.~\ref{fig:le_nle_fig}(B-D) where nodes 5, 11, and 12 appear in local neighborhood of node 3 but node 9 is eliminated. Because nodes 5, 11, and 12, despite of being different class, are correlated and densely connected with node 3 while MI between nodes 3 and 9 is $0.0$ meaning that they are irrelevant to each other. But non-local neighborhood of node 3 consists of nodes 8, 10, and 13 as they are highly correlated in their latent space. Moreover, for node 1, local structure (e.g. clique) can be captured through local neighborhood but it is lost in non-local neighborhood. 

\subsection{Attentive Aggregation, AA:}
In feature aggregation, all nodes are not likely to be equally important to a particular node. Following the idea of GAT\cite{velivckovic2017graph}, we compute attention coefficient, $e_{uv}$ for nodes $v \in \mathcal{N}_{u}$ where $\mathcal{N}_{u}$ is some neighborhood of node $u$. Afterward, the normalized attention coefficients, $a_{uv}$ are used to compute a linear combination of the features of respective nodes. The resulting feature vector after applying non-linearity, $h_{u}$ would be considered as an aggregated feature embedding of node $u$,
\begin{equation}
\begin{split}
    &e_{uv} = \alpha(\widetilde{a}^{T}[W\mathcal{Z}_{u}||W\mathcal{Z}_{v}]) \\
    &a_{uv} = \frac{exp(e_{uv})}{\sum_{\bar{v} \in \mathcal{N}_{u}} exp(e_{u\bar{v}})} \\
    &h_{u} = \alpha(\sum_{\forall v \in \mathcal{N}_{u}}a_{uv} \cdot X_{v})
\end{split}
\label{eq:aa}
\end{equation}
where $\widetilde{a}$ and $W$ are two learnable parameters which are shareable among all node-pairs. $T$ and $||$ represent transposition and concatenation operation respectively while $\alpha$ is used to incorporate nonlinearity. $\mathcal{Z}_{u}$ and $\mathcal{Z}_{v}$ denote the self-embedding of node $u$ and $v$. The more correlated two nodes are in the self-embedding space, the larger the value of the coefficient should be.

\subsection{Bi-level Aggregation}
\subsubsection{Local Aggregation, LA} We perform attentive aggregation (AA) on local neighborhood $\mathcal{N_{L}}$ that enable our model to accumulate features from relevant densely connected nodes: 
\begin{equation}
    \mathcal{H_{L}}(u) = ReLU(W_{\mathcal{L}} \cdot AA(\{ X_{v},\forall v \in \mathcal{N_{L}}(u) \}))
    \label{eqn:LA}
\end{equation}
\subsubsection{Non-Local Agrgegation, nLA}
We employ attentive aggregation (AA) on non-local neighborhood $\mathcal{N_{\widetilde{L}}}$ to aggregate important features from distant but informative nodes:
\begin{equation}
    \mathcal{H_{\widetilde{L}}}(u) = ReLU(W_{\mathcal{\widetilde{L}}} \cdot AA(\{ X_{v},\forall v \in \mathcal{N_{\widetilde{L}}}(u) \}))
    \label{eqn:nLA}
\end{equation}
After that, both local and non-local embeddings of nodes are combined and multiplied with weights to compute final embeddings as follows:
\begin{equation}
    \mathcal{H_{F}}(u) = ReLU(W_{\mathcal{F}} \cdot (X_{u} || \mathcal{H_{L}}(u) || \mathcal{H_{\widetilde{L}}}(u)))
    \label{eqn:final}
\end{equation}

Where, Ws indicate learnable weights for corresponding equations, $||$ is concatenation and AA means Attentive Aggregation. $\mathcal{H_{F}}$ indicates the desired final embeddings of nodes. Eq.~\ref{eqn:final} indicates that our model has the expressive potential to decide whether to use local or non-local information or both based on input graphs. In Fig.~\ref{fig:model_diagram}, our model trains MIE and compute SEs by optimizing the estimation loss in Eq.~\ref{eq:mi_loss} at first stage. Then in the next stage, it updates the parameters of the bi-level aggregation module by optimizing the classification cross-entropy loss.
\subsection{Time Complexity:} Our model takes $O(|E|)$ to find the optimal communities. Whereas it needs $O(|L|*|V|)$ for each clustering iteration. We utilize label data into the initialization step for the faster convergence of clustering.

\begin{table*}[t]
\centering
\scalebox{0.9}{
\begin{tabular}{c|ccccccc||ccc} \hline
\multicolumn{8}{c||}{\textbf{Disassortative}} & \multicolumn{3}{c}{\textbf{Assortative}} \\ \hline 
\textbf{Models} & \textit{Texas} & \textit{Cornell}  & \textit{Wisconsin} & \textit{Washington}  & \textit{Squirrel} & \textit{Actor} & \textit{Chameleon}  & \textit{Citeser} & \textit{Pubmed}  & \textit{Cora}     \\ \hline
GCN~\cite{kipf2016semi}             & 56.9     & 53.4                & 58.9  & 59.5                      & 35.3              & 29.8            & 56.1                      & 75.8             & 87.7 & 87.2         \\ 
GAT~\cite{velivckovic2017graph}         & 56.7    & 55.4                 & 58.8      & 57.3          & 33.1              & 27.8            & 52.6         & 76.6             & 87.0        & \underline{87.5}             \\ 
GraphSAGE~\cite{hamilton2017inductive}        & 82.6          &74.2             &78.1               &78.3              & 39.5              & 32.3      & 62.8             & 76.2                & 88.7          &86.8     \\ 

GEOM-GCN*~\cite{pei2020geom}            &73.5     & 75.4           & 80.3                & -                 &46.0          &34.5         &68.0             & \underline{80.6}            & \underline{90.7}      & 87.0   \\
NL-GNN*~\cite{liu2020non}               & \underline{85.4}        & \underline{84.9}                & \underline{87.3}      & -      & \underline{59.0}              & \underline{37.9}       & \underline{70.1}             & -             & -                & -            \\
\textbf{LnL-GNN(Ours)}            & \textbf{91.2}     & \textbf{90.3}               & \textbf{91.4} &       \textbf{90.8}          &\textbf{62.3} & \textbf{40.8}          & \textbf{72.2}                 & \textbf{81.4}                & \textbf{91.3}              & \textbf{89.5} \\ \hline
MLP-Raw            &81.1     &82.2            &83.1                &\underline{81.8}     & 32.3          & 34.4     &49.2          & 70.3           & 86.7     & 73.6      \\
MLP-Mean             &63.7     &47.4           &55.1                &57.6          & 48.6          & 25.4    &63.9          & 73.7           & 85.4         & 84.7     \\\hline
\end{tabular}
}
\caption{Mean Classification Accuracy (Percent): Results of GEOM-GCN* \& NL-GNN* obtained from~\cite{pei2020geom} \& \cite{liu2020non} respectively. (\textbf{Best}, \underline{2nd Best}). It is to be noted that the standard deviation of accuracies of LnL-GNN over 10 runs varies from 0.5 -1.5 in assortative graphs while 2.5 - 3.5 in dissassortive graphs. '-' denotes the results are not publicly available.}
\label{tab:performance_comparison}
\end{table*}

\section{Experiments}
We validate our model by comparing its performance on transductive node classification tasks with recent popular baseline models. Besides, further analysis of local and non-local aggregation demonstrates the efficacy of our model.

\subsection{Dataset Description:}
We conduct our experiments on ten open graph datasets. They are categorized into two groups as assortative and disassortative based on their homophily ratio(HR). An overview summary of the datasets are given in Table~\ref{tab:dataset_description} where assortative datasets are three citation networks: Cora, Citeser, and Pubmed and disassortative datasets consist of Wikipedia networks: Chameleon and Squirrel, co-occurrence network: Actor and  webKB\footnote{http://www.cs.cmu.edu/afs/cs.cmu.edu/project/theo-11/www/wwkb/} datasets: Cornell, Texas, Wisconsin, Washington~\cite{pei2020geom}.

\subsection{Experimental Setup:}
We tune hyperparameters for our models as follows: i) the number of layers in MLP $\in \{2\}$, ii) the dimension of self-embedding $\in \{128\}$, iii) dropout rate $\in \{0.25\}$, iv) learning rate $\in \{0.01\}$, v) weight decay $\in \{0.0001\}$, vi) the number of attention head $\in \{5\}$, vii) the number of neighbors sampled from non-local neighborhood $\in \{128\}$, viii) the number of aggregation layers $\in \{2\}$, ix) the dimension of hidden representations $\in \{128\}$, and x) size of mini-batches $\in\{2048\}$ (Squirrel, Actor and Pubmed). We use SGD optimizer with momentum value of 0.9. We randomly split nodes of each class into 60\%, 20\%, 20\% for training, validation, and testing for each dataset. We run experiments 10 times with different random splits and report the average accuracy with the patience of 100 epochs over 10 runs. However, we run GCN~\cite{kipf2016semi}, GAT~\cite{velivckovic2017graph}, GraphSAGE~\cite{hamilton2017inductive} on benchmarking datasets with their mentioned parameter settings.

\subsection{Comparison with Baselines:}
Initially, we conduct experiments on the simplest baselines: MLP-Raw that makes predictions solely based on node raw features and MLP-Mean on mean aggregated features of $1$-hop neighbor nodes. 
Table~\ref{tab:performance_comparison} shows that the raw features (MLP-Raw) are more influential in classification than mean aggregated features of 1-hop neighbor nodes (MLP-Mean) on Texas, Cornell, Washington, Wisconsin, Pubmed, and Actor datasets. But for the rest of the datasets, the opposite holds. Thus, we compute self-embeddings from mean aggregated features of 1-hop neighbors instead of target nodes' (ego) features on Cora, Citeseer, Chameleon, and Squirrel datasets. In contrast, node raw features without any aggregation are embedded into SEs in the rest of the datasets. In Table~\ref{tab:performance_comparison}, we can see that our model, named LnL-GNN, consistently outperforms all baseline approaches on both disassortative and assortative graph datasets. In case of disassortative graphs where distant informative nodes provide more useful features than local neighbors, local-aggregation based approaches such as GCN, GAT and GraphSAGE underperform because they fail to capture important features from distant nodes due to their limited receptive fields. It is to be noted that GraphSAGE performs better than GCN and GAT because of keeping ego (target node) and aggregated neighbor-embeddings separate. However, it performs on par with GEOM-GCN and MLP variants.
Despite GEOM-GCN's geometric aggregation scheme, it underperforms due to some concerning limitations: a) use of non-task-specific pre-trained methods as mapping function to the embedded space; b) highly constrained mapping of node features to 2-D latent space (as specified in their experimental section) that is not enough to preserve discriminative properties of nodes, etc. NL-GNN shows promising performance, but its attention-guided sorting with a single calibration vector is incompetent to push distant but informative nodes together. Whereas the key factors behind the success of our model in both assortative and disassortative graphs are: i) MI-based clustering can effectively group all distant correlated nodes into the same clusters resulting in informative non-local neighborhoods that effectively leverage relevant information into non-local embeddings; ii) the local neighborhood module has the ability to distinguish densely inter-connected correlated nodes from noisy ones to form data-specific more informative receptive field, and iii) Eq.~\ref{eqn:final} keeps ego (target node), local and non-local embeddings separate without mixing them which steers the model to achieve its best generalizing abilities through weighing them differently.
\subsection{Analysis of Local Neighborhood:}
\begin{figure}[t]
\centering

\centering
\includegraphics[width=\columnwidth]{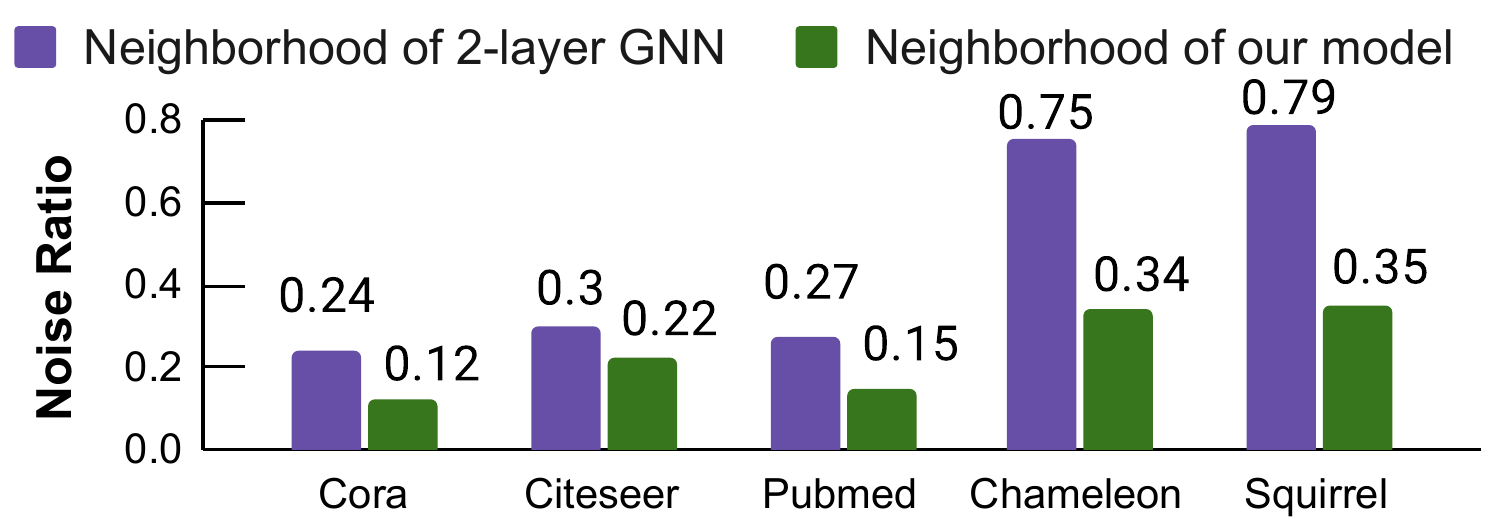}
\caption{Effectiveness of Local Neighborhood: Comparisons of noise ratio between 2-hop neighborhood of existing GNNs  and local neighborhood of our model.}
\label{fig:lo_nr}
\end{figure}

The main motivation of MI-guided community-based local neighborhoods in our model is to filter out irrelevant/noisy nodes from informative neighbors. To validate the effectiveness of our model, we show the comparisons of noise ratio~\cite{hou2019measuring} defined as $NR(G) = \frac{1}{|V|}\sum_{v \in V}\frac{\Bar{L^{2}_{v}}}{L^{2}_{v}}$ where $L^{2}_{v}$ and $\Bar{L^{2}_{v}}$ denote the number of nodes within 2-hop away and that have different class labels from $v$ respectively. In Fig.~\ref{fig:lo_nr}, the noise ratio of the 2-hop neighborhood of our model is significantly smaller than that of existing GNNs, where the smaller the value is, the less susceptible to noise the model is. These results indicate that our model can learn distinguishable representations by leveraging important features without being \textit{'washed out'} by incorporating noise while existing GNNs suffer from over-smoothing problems due to the high noise ratio.

\begin{figure}[h]
\centering
\includegraphics[width=\columnwidth]{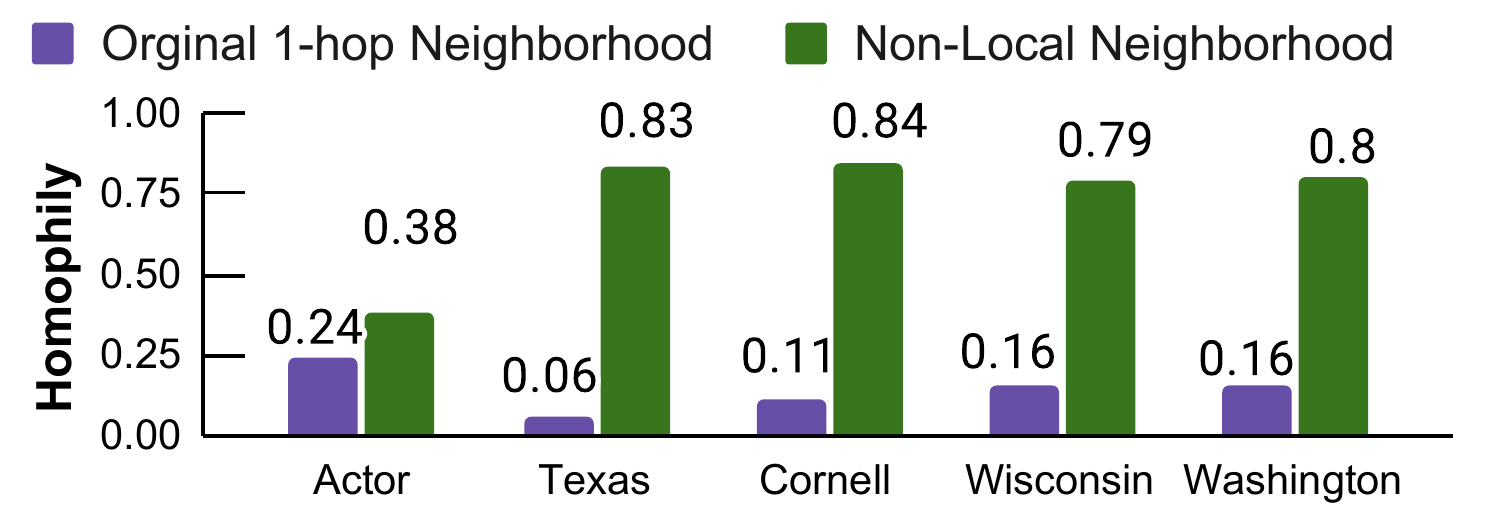}
\caption{Effectiveness of Non-Local Neighborhood: Comparisons of homophily ratio between 1-hop neighborhood on input graph and non-local neighborhood of our model.}
\label{fig:nl_hr}
\end{figure}

\subsection{Analysis of Non-Local Neighborhood:}
To leverage important features from distant informative nodes, they should appear in each other's non-local neighborhoods. In other words, the homophily ratio of the non-local neighborhood should be larger than the original $1$-hop neighborhood. We compute the homophily ratio of the non-local neighborhood and exhibit comparisons in Fig.~\ref{fig:nl_hr}. In Fig.~\ref{fig:nl_hr}, we can see that the homophily ratio of the non-local neighborhood is much larger than the original $1$-hop neighborhood for all five datasets. These comparisons indicate that non-local aggregation on MI-based non-local neighborhood can effectively bring more similar nodes together into the same cluster. As a result, it can utilize important features from distant similar nodes to learn hidden representations. 


\begin{table}[h]
\small
\scalebox{0.95}{
\begin{tabular}{c|c|c|c}
\hline
Dataset &
  \begin{tabular}[c]{@{}c@{}}Local \\ Aggregation\end{tabular} &
  \begin{tabular}[c]{@{}c@{}}Non-Local \\ Aggregation\end{tabular} &
  \multicolumn{1}{c}{\begin{tabular}[c]{@{}c@{}}Bi-level Aggregation\end{tabular}} \\ \hline
Texas      & 65.70  & 87.07 & 91.2 \\ 
Cornell    & 60.45 & 86.28 & 90.3 \\ 
Wisconsin  & 64.10& 88.12 & 91.4 \\
Washington & 63.99  & 87.72 &  90.8 \\ 
 Squirrel  & 50.67 & 58.31 &62.3  \\ 
Actor      & 33.23 & 38.12 &40.8  \\ 
Chameleon & 65.12 &  70.37 & 72.2 \\ \hline 
Citeser  & 79.80 & 76.40&81.4   \\
 Pubmed    & 90.34 & 88.96 &91.3  \\
 Cora  & 88.15 & 86.64 &  89.5\\ \hline
\end{tabular}
\caption{Performance of separate Local and Non-Local Aggregation module compared to Bi-level Aggregation module.}
\label{tab:ablation_local_non_local}
}
\end{table}

\subsection{Ablation Study on Local and Non-Local Aggregation} In Table~\ref{tab:ablation_local_non_local}, only local aggregation underperform in disassortative graphs because their local neighbors are less relevant and informative. In contrast, it performs well for the assortative graphs. However, the promising performance of non-local aggregation in disassortative graphs demonstrates the effectiveness of non-local neighborhood. But non-local aggregation underperforms local aggregation in assortative graphs. It is clear that densely connected relevant neighbors are more important than distant relevant nodes in assortative graphs. However, Bi-level aggregation consistently outperforms local and non-local aggregation in both assortative and disassortative graphs because it takes the advantages of of both local and non-local aggregation that leads to superior performance.

\section{Conclusions}
In this work, we exploit mutual information (MI) to address two major drawbacks of existing GNNs - important features being \textit{'washed out'} in local aggregation and the lack of ability to capture long-range non-linear dependencies among nodes. Besides, we leverage self-supervision learning to train MI estimator with few label data. Our proposed local-aggregation module can draw data-specific local neighborhood to compute local embeddings by filtering out irrelevant nodes, that enables our model to tackle the over-smoothing problem to some extent. Further, our non-local aggregation module utilizes a clustering technique with MI to capture non-local dependencies where distant but informative nodes are grouped into the same cluster to compute non-local embeddings. Superior results from extensive experiments prove the efficacy of our model on node classification tasks and its effectiveness in overcoming the two drawbacks mentioned above. 

\section{Acknowledgements}
This project is supported by Independent University Bangladesh and ICT Division of Bangladesh Government.

\bibliographystyle{IEEEtran}
\bibliography{reference}

\end{document}